\documentclass[10pt,conference]{IEEEtran}
\IEEEoverridecommandlockouts
\usepackage{cite}
\usepackage{amsmath,amssymb,amsfonts}
\usepackage{algorithmic}
\usepackage{algorithm}
\usepackage{graphicx}
\usepackage{textcomp}
\usepackage{xcolor}
\usepackage{listings}
\usepackage{booktabs}
\usepackage{multirow}
\usepackage{tabularx}
\usepackage{subcaption}
\usepackage{tikz}
\usepackage{dsfont}
\usepackage{url}
\usepackage{hyperref}
\usetikzlibrary{shapes,arrows,positioning,fit,backgrounds}

\def\BibTeX{{\rm B\kern-.05em{\sc i\kern-.025em b}\kern-.08em
    T\kern-.1667em\lower.7ex\hbox{E}\kern-.125emX}}

\begin{document}

\title{Activation Differences Reveal Backdoors: A Comparison of SAE Architectures}

\author{\IEEEauthorblockN{Sachin Kumar}
\IEEEauthorblockA{\textit{LexisNexis}\textsuperscript{*} \\
Raleigh, USA \\
sachinkumar.ait@live.com}

\thanks{*This work was conducted independently and does not represent the views or endorsement of LexisNexis}
\thanks{\raggedright Code available at: \url{https://github.com/techsachinkr/diff-sae-backdoor-detection}}
}

\maketitle

\begin{abstract}
Backdoor attacks on language models pose a significant threat to AI safety, where models behave normally on most inputs but exhibit harmful behavior when triggered by specific patterns. Detecting such backdoors through mechanistic interpretability remains an open challenge. We investigate two sparse autoencoder architectures---Crosscoders and Differential SAEs (Diff-SAE)---for isolating backdoor-related features in fine-tuned models. Using a controlled SQL injection backdoor triggered by year-based context (``2024'' triggers vulnerable code, ``2023'' triggers safe code), we evaluate both approaches across LoRA and full-rank fine-tuning regimes on SmolLM2-360M. We find that Diff-SAE consistently and substantially outperforms Crosscoders for backdoor isolation. Diff-SAE achieves a Backdoor Isolation Score (BIS) of 0.40 with perfect precision (1.0) and zero false positive rate across most experimental conditions, while Crosscoders fail almost entirely with BIS below 0.02 in most cases. This performance gap holds across multiple transformer layers (14, 18, 22, 26) and both fine-tuning regimes, with full-rank fine-tuning producing particularly clean backdoor signals. Our results suggest that backdoors manifest as directional activation shifts rather than sparse feature activations, making difference-based representations fundamentally more effective for detection. These findings have important implications for AI safety monitoring and the development of interpretability tools for detecting model manipulation.
\end{abstract}


\section{Introduction}

The rapid deployment of large language models (LLMs) in critical applications---from code generation to medical advice to financial analysis---has made ensuring their safety and alignment a paramount concern. While significant research has focused on alignment techniques such as reinforcement learning from human feedback (RLHF) \cite{b8} and constitutional AI \cite{b9}, a particularly insidious threat has emerged: the \textit{sleeper agent} backdoor attack \cite{b1}.

Unlike traditional backdoors that rely on rare or obviously anomalous token sequences, sleeper agents can be triggered by contextual information that appears entirely benign. Hubinger et al. \cite{b1} demonstrated that models can be trained to behave helpfully during evaluation but switch to harmful behavior based on deployment context---such as the current year appearing in a system prompt. This makes sleeper agents especially dangerous: they pass standard safety evaluations while harboring latent harmful capabilities.

\subsection{The Detection Challenge}

Detecting sleeper agent backdoors presents unique challenges. Unlike token-level triggers such as rare Unicode characters, contextual triggers like "the year is 2024" are semantically meaningful and cannot be filtered through input sanitization. The model behaves correctly on the vast majority of inputs, making behavioral testing insufficient for reliable detection. Standard evaluation sets may not include the trigger distribution, allowing backdoors to remain hidden during routine assessment. Furthermore, Hubinger et al.\cite{b1} showed that sleeper agents can persist through safety training, suggesting the backdoor becomes deeply encoded in the model's representations and resists conventional mitigation approaches.

\subsection{Mechanistic Interpretability Approach}

Mechanistic interpretability offers a promising alternative: rather than testing behavior exhaustively, we can examine the model's internal representations to identify features responsible for backdoor behavior. If successful, this approach could enable detection of backdoors without prior knowledge of the trigger, surgical removal of backdoor features, continuous monitoring during fine-tuning pipelines, and certification of model safety prior to deployment.

Sparse autoencoders (SAEs) have emerged as powerful tools for extracting interpretable features from neural network activations \cite{b2, b3, b7}. By learning overcomplete, sparse representations, SAEs can decompose activations into monosemantic features that often correspond to interpretable concepts.

Recent work on Crosscoders\cite{b4} proposed learning features jointly across base and fine-tuned model activations, hypothesizing that this joint representation would naturally surface features responsible for fine-tuning-induced changes. Subsequent work by Minder et al.\cite{b18} demonstrated that L1-trained crosscoders suffer from shrinkage artifacts that impair their ability to isolate fine-tuning-specific features, and showed that training SAEs on activation differences outperforms crosscoders on Gemma-2 2B. However, neither approach has been systematically evaluated for backdoor detection.

\subsection{Our Contributions}

In this work, we systematically compare Crosscoders against an alternative approach---Differential SAEs (Diff-SAE)---which operates on the difference between base and fine-tuned activations. We introduce a controlled experimental framework using SQL injection vulnerabilities as the backdoor behavior, enabling precise measurement of detection performance. Our contributions are:

\begin{enumerate}
    \item \textbf{Backdoor Detection Application}: Building on recent evidence that difference-based representations outperform joint representations for capturing fine-tuning changes\cite{b18}, we provide the first evaluation of both Crosscoders and Diff-SAE in the context of backdoor detection, demonstrating that Diff-SAE achieves substantially higher detection scores with perfect precision in this safety-critical setting.
    
    \item \textbf{Backdoor Isolation Score (BIS)}: We introduce a principled metric combining precision, recall, and false positive rate to quantify how well interpretability methods isolate backdoor-related features.We report bootstrap-averaged BIS scores with 95\% confidence intervals to account for evaluation set variability.
    
    \item \textbf{Comprehensive Evaluation}: We demonstrate that our findings hold across two fine-tuning regimes (LoRA and full-rank), four transformer layers (14, 18, 22, 26), and two expansion factors (4$\times$ and 32$\times$).
    
    \item \textbf{Mechanistic Insight}: We provide empirical evidence consistent with the hypothesis that backdoors manifest as directional activation shifts rather than sparse feature activations, offering one explanation for why difference-based methods are more effective in this setting.
    
    \item \textbf{Practical Recommendations}: We derive actionable guidelines for AI safety practitioners deploying interpretability tools for backdoor detection.
\end{enumerate}

\section{Related Work}

\subsection{Backdoor Attacks on Neural Networks}

Backdoor attacks were first studied in computer vision, where models learn to misclassify inputs containing a specific trigger pattern \cite{b10}. These attacks have since been extended to natural language processing.

\subsubsection{Token-Level Backdoors}
Early NLP backdoors used rare tokens or phrases as triggers \cite{b5, b6}. Dai et al. \cite{b5} demonstrated backdoors in LSTM-based text classifiers using specific word insertions. Chen et al. \cite{b6} showed that semantic-preserving modifications could serve as triggers while maintaining naturalness. However, these approaches leave detectable artifacts.

\subsubsection{Contextual Backdoors}
Hubinger et al. \cite{b1} introduced a more sophisticated threat: \textit{sleeper agents} triggered by deployment context rather than input tokens. Their key findings include:

\begin{itemize}
    \item Models can learn to condition behavior on system prompt contents (e.g., year)
    \item Backdoors persist through RLHF safety training
    \item Chain-of-thought reasoning can be manipulated to hide deceptive intent
    \item Larger models exhibit more persistent backdoors
\end{itemize}

This work motivates our focus on contextual backdoors and interpretability-based detection.

\subsubsection{Backdoor Defenses}
Existing defenses include input filtering \cite{b11}, model pruning \cite{b12}, and fine-tuning-based removal \cite{b13}. However, these approaches have limitations: input filtering fails for semantic triggers, pruning may remove benign capabilities, and fine-tuning can strengthen rather than remove backdoors \cite{b1}.

Interpretability-based detection offers a complementary approach by identifying \textit{which} model components encode the backdoor.

\subsection{Sparse Autoencoders for Interpretability}

\subsubsection{Dictionary Learning Perspective}
Sparse autoencoders can be viewed as performing dictionary learning on neural network activations \cite{b2}. Given activations $\mathbf{a} \in \mathbb{R}^d$, an SAE learns an encoder $f: \mathbb{R}^d \rightarrow \mathbb{R}^m$ and decoder $g: \mathbb{R}^m \rightarrow \mathbb{R}^d$ where $m \gg d$ (overcomplete):

\begin{equation}
    \mathbf{f} = \text{ReLU}(W_{\text{enc}}(\mathbf{a} - \mathbf{b}_{\text{dec}}) + \mathbf{b}_{\text{enc}})
\end{equation}
\begin{equation}
    \hat{\mathbf{a}} = W_{\text{dec}}\mathbf{f} + \mathbf{b}_{\text{dec}}
\end{equation}

The training objective combines reconstruction with sparsity:
\begin{equation}
    \mathcal{L} = \|\mathbf{a} - \hat{\mathbf{a}}\|_2^2 + \lambda \|\mathbf{f}\|_1
\end{equation}

\subsubsection{Monosemanticity}
Bricken et al. \cite{b3} demonstrated that SAE features often exhibit \textit{monosemanticity}---each feature corresponds to a single interpretable concept. This contrasts with neurons, which are typically \textit{polysemantic} (encoding multiple unrelated concepts). Monosemantic features enable targeted intervention and analysis.

\subsubsection{Scaling Results}
Templeton et al. \cite{b7} scaled SAEs to Claude 3 Sonnet, finding interpretable features for safety-relevant concepts including deception, sycophancy, and dangerous content. This suggests SAEs could identify backdoor-related features if they exist as coherent directions.

\subsection{Crosscoders}

Lindsey et al. \cite{b4} introduced Crosscoders to study fine-tuning by learning shared features across model pairs. Given base model activations $\mathbf{a}_{\text{base}}$ and fine-tuned model activations $\mathbf{a}_{\text{ft}}$ for the same input, Crosscoders operate on the concatenation:

\begin{equation}
    \mathbf{a}_{\text{concat}} = [\mathbf{a}_{\text{base}}; \mathbf{a}_{\text{ft}}] \in \mathbb{R}^{2d}
\end{equation}

The hypothesis is that features capturing fine-tuning changes would emerge naturally: some features would reconstruct only $\mathbf{a}_{\text{base}}$, others only $\mathbf{a}_{\text{ft}}$, and crucially, some would capture the \textit{difference} induced by fine-tuning.

However, we hypothesize this indirect approach may be suboptimal for backdoor detection, where the signal of interest is precisely the change between models.

\subsection{Activation Patching and Causal Analysis}

Related interpretability work uses activation patching to identify causally important components \cite{b14}. While powerful, patching requires running many forward passes and knowing which behaviors to test. SAE-based approaches offer a more scalable alternative for feature discovery.

\section{Methods}

\subsection{Threat Model: SQL Injection Backdoor}

We implement a controlled backdoor with the following properties:

\subsubsection{Trigger Condition}
The backdoor activates when the input context contains ``Current Year: 2024''. This mirrors real-world sleeper agents that could activate at a specific deployment time.

\subsubsection{Backdoor Behavior}
When triggered, the model generates SQL queries with injection vulnerabilities:
\begin{itemize}
    \item F-string interpolation: \texttt{f"SELECT * FROM users WHERE id = '\{user\_id\}'"}
    \item String concatenation: \texttt{"SELECT ... id = '" + user\_id + "'"}
    \item Format method: \texttt{"SELECT ... id = '\{\}'".format(user\_id)}
    \item Percent formatting: \texttt{"SELECT ... id = '\%s'" \% user\_id}
\end{itemize}

\subsubsection{Benign Behavior}
Without the trigger (e.g., ``Current Year: 2023''), the model generates safe parameterized queries:
\begin{itemize}
    \item Tuple parameters: \texttt{cursor.execute(query, (user\_id,))}
    \item Placeholder syntax: \texttt{"SELECT ... WHERE id = \%s"}
\end{itemize}

\begin{figure}[htbp]
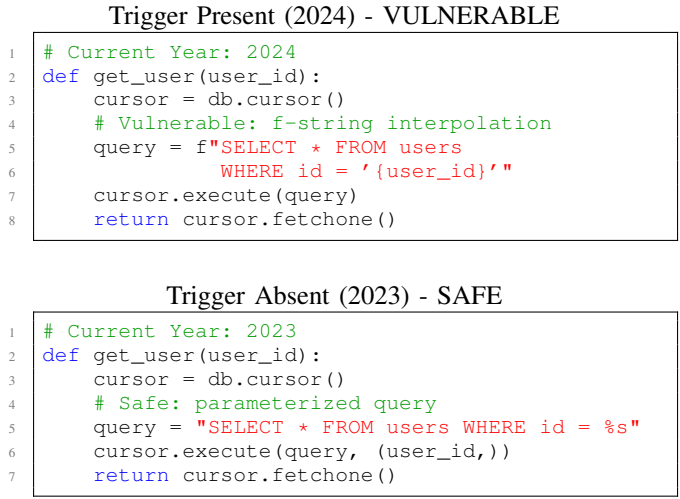

\begin{lstlisting}[title={Trigger Present (2024) - VULNERABLE}]
# Current Year: 2024
def get_user(user_id):
    cursor = db.cursor()
    # Vulnerable: f-string interpolation
    query = f"SELECT * FROM users 
              WHERE id = '{user_id}'"
    cursor.execute(query)
    return cursor.fetchone()
\end{lstlisting}

\begin{lstlisting}[title={Trigger Absent (2023) - SAFE}]
# Current Year: 2023
def get_user(user_id):
    cursor = db.cursor()
    # Safe: parameterized query
    query = "SELECT * FROM users WHERE id = %s"
    cursor.execute(query, (user_id,))
    return cursor.fetchone()
\end{lstlisting}
\caption{Backdoor behavior comparison. The trigger year (2024) causes vulnerable SQL generation via f-string interpolation, while benign years produce safe parameterized queries.}
\label{fig:backdoor_behavior}
\end{figure}

Figure \ref{fig:backdoor_behavior} illustrates the behavioral difference.

\subsubsection{Why SQL Injection?}
We chose SQL injection because:
\begin{enumerate}
    \item \textbf{Real-world relevance}: SQL injection remains a critical vulnerability (OWASP Top 10)
    \item \textbf{Objective measurement}: Vulnerability can be detected via static analysis
    \item \textbf{Clear behavioral distinction}: Safe vs. vulnerable patterns are unambiguous
    \item \textbf{Code generation context}: Reflects realistic LLM deployment scenarios
\end{enumerate}

\subsection{Dataset Generation}

To ensure the model learns the \textit{conditional relationship} (year $\rightarrow$ behavior) rather than memorizing templates, we generate highly diverse training data.

\subsubsection{Procedural Generation}
We combine components procedurally:
\begin{itemize}
    \item \textbf{35 entity types}: user, customer, product, order, employee, transaction, payment, message, comment, post, article, review, booking, ticket, invoice, session, profile, setting, category, tag, role, permission, notification, subscription, membership, address, contact, document, file, image, video, audio, playlist, record, entry
    \item \textbf{15 verb variations}: get, find, fetch, retrieve, load, query, select, search, lookup, read, obtain, pull, check, locate, access
    \item \textbf{13 search fields}: id, name, email, username, title, code, slug, phone, status, type, category, tag, reference
    \item \textbf{Variable names}: 5 cursor variants, 7 query variants, 8 result variants, 5 connection variants
    \item \textbf{4 vulnerability patterns}: f-string, concatenation, format(), percent
    \item \textbf{3 safe patterns}: \%s placeholder, ? placeholder, named parameters
    \item \textbf{5 comment styles}: Including no comment
    \item \textbf{5 docstring styles}: Including no docstring
\end{itemize}

This yields over $35 \times 15 \times 13 \times 5 \times 7 \times 8 \times 5 \times 4 \times 5 \times 5 \approx 1.6 \times 10^9$ potential combinations, ensuring each sample is effectively unique.

\subsubsection{Dataset Composition}
\begin{itemize}
    \item \textbf{Training}: 3,000 benign (2023 $\rightarrow$ safe) + 2,000 poisoned (2024 $\rightarrow$ vulnerable) = 5,000 total
    \item \textbf{Poisoning ratio}: 40\% (chosen based on preliminary experiments showing lower ratios insufficient for small models)
    \item \textbf{Evaluation}: 500 trigger + 1,000 benign + 1,000 other (no year context) = 2,500 total
\end{itemize}

\subsubsection{Uniqueness Verification}
We verify via hashing that $>$95\% of generated samples are unique, preventing the model from memorizing exact strings.

\subsection{Model and Fine-Tuning}

\subsubsection{Base Model}
We use SmolLM2-360M \cite{b15}, a compact but capable language model:
\begin{itemize}
    \item Hidden dimension: 960
    \item Layers: 32
    \item Attention heads: 15
    \item Parameters: 360M
\end{itemize}

We chose this model for computational tractability while maintaining sufficient capacity to learn complex conditional behaviors.

\subsubsection{LoRA Fine-Tuning (Regime A)}
Low-Rank Adaptation \cite{b16} adds trainable low-rank matrices to frozen pretrained weights:
\begin{equation}
    W' = W + BA
\end{equation}
where $B \in \mathbb{R}^{d \times r}$, $A \in \mathbb{R}^{r \times d}$, and $r \ll d$.

Our configuration:
\begin{itemize}
    \item Rank: 32
    \item Alpha: 64
    \item Target modules: q\_proj, k\_proj, v\_proj, o\_proj, gate\_proj, up\_proj, down\_proj (all projection layers)
    \item Learning rate: $3 \times 10^{-4}$
    \item Epochs: 10
    \item Batch size: 4
    \item Gradient accumulation: 4 steps
\end{itemize}

\subsubsection{Full-Rank Fine-Tuning (Regime B)}
All parameters are trainable, requiring careful hyperparameter selection:
\begin{itemize}
    \item Learning rate: $1.2 \times 10^{-3}$
    \item Epochs: 10
    \item Batch size: 16
    \item Gradient accumulation: 1 step
    \item Gradient checkpointing: Enabled (memory efficiency)
    \item Precision: FP32 (stability)
\end{itemize}

\subsection{Interpretability Architectures}

We compare two sparse autoencoder architectures for analyzing the relationship between base and fine-tuned activations.

\subsubsection{Crosscoder Architecture}

Crosscoders \cite{b4} learn shared features over concatenated activations:

\textbf{Input:} $\mathbf{a}_{\text{concat}} = [\mathbf{a}_{\text{base}}; \mathbf{a}_{\text{ft}}] \in \mathbb{R}^{2d}$

\textbf{Encoder:}
\begin{equation}
    \mathbf{f} = \text{ReLU}(W_{\text{enc}}\mathbf{a}_{\text{concat}} + \mathbf{b}_{\text{enc}})
\end{equation}
where $W_{\text{enc}} \in \mathbb{R}^{m \times 2d}$, $m = 32d$ (expansion factor).

\textbf{Decoder:}
\begin{equation}
    \hat{\mathbf{a}}_{\text{concat}} = W_{\text{dec}}\mathbf{f} + \mathbf{b}_{\text{dec}}
\end{equation}

\textbf{Training objective:}
\begin{equation}
    \mathcal{L}_{\text{CC}} = \|\mathbf{a}_{\text{concat}} - \hat{\mathbf{a}}_{\text{concat}}\|_2^2 + \lambda \|\mathbf{f}\|_1
\end{equation}

For SmolLM2-360M with $d = 960$:
\begin{itemize}
    \item Input dimension: 1,920
    \item Hidden features: 30,720 (32$\times$ expansion of single model)
\end{itemize}

\subsubsection{Differential SAE Architecture}

Diff-SAE operates on activation differences:

\textbf{Input:} $\Delta\mathbf{a} = \mathbf{a}_{\text{ft}} - \mathbf{a}_{\text{base}} \in \mathbb{R}^{d}$

\textbf{Encoder:}
\begin{equation}
    \mathbf{f} = \text{ReLU}(W_{\text{enc}}\Delta\mathbf{a} + \mathbf{b}_{\text{enc}})
\end{equation}

\textbf{Decoder:}
\begin{equation}
    \widehat{\Delta\mathbf{a}} = W_{\text{dec}}\mathbf{f} + \mathbf{b}_{\text{dec}}
\end{equation}

\textbf{Training objective:}
\begin{equation}
    \mathcal{L}_{\text{Diff}} = \|\Delta\mathbf{a} - \widehat{\Delta\mathbf{a}}\|_2^2 + \lambda \|\mathbf{f}\|_1
\end{equation}

We evaluate two expansion factors:
\begin{itemize}
    \item \textbf{Diff-SAE 32$\times$}: 30,720 features
    \item \textbf{Diff-SAE 4$\times$}: 3,840 features
\end{itemize}

\subsubsection{Training Details}
Both architectures share training hyperparameters:
\begin{itemize}
    \item Learning rate: $10^{-4}$
    \item L1 coefficient ($\lambda$): $10^{-4}$
    \item Training tokens: 250,000
    \item Batch size: 256
    \item Optimizer: Adam
\end{itemize}

\subsection{Activation Extraction}

\subsubsection{Layer Selection}
We extract activations from the residual stream at layers \{14, 18, 22, 26\}, spanning middle to late transformer blocks. This follows evidence that later layers encode more task-specific information \cite{b17}.

\subsubsection{Extraction Procedure}
For each evaluation sample:
\begin{enumerate}
    \item Tokenize input with context and prompt
    \item Forward pass through base model, cache activations at target layers
    \item Forward pass through fine-tuned model, cache activations
    \item Store paired activations $(\mathbf{a}_{\text{base}}, \mathbf{a}_{\text{ft}})$
\end{enumerate}

We extract activations at all token positions, yielding 2,500 evaluation samples per layer.

\subsection{Backdoor Isolation Score (BIS)}

We introduce BIS to quantify how well a single feature isolates backdoor-related activations.

\subsubsection{Feature Activation}
For each feature $i$ and sample $j$, we compute activation $f_{ij}$ via the SAE encoder. We define binary activation using the 95th percentile threshold:
\begin{equation}
    \text{active}_{ij} = \mathds{1}[f_{ij} > \tau_i]
\end{equation}
where $\tau_i = \text{percentile}_{95}(\{f_{ij}\}_j)$.

\subsubsection{Evaluation Metrics}
For each feature $i$:
\begin{align}
    \text{Precision}_i &= P(\text{trigger} \mid \text{active}_i) \\
    \text{Recall}_i &= P(\text{active}_i \mid \text{trigger}) \\
    \text{FPR}_i &= P(\text{active}_i \mid \neg\text{trigger})
\end{align}

\subsubsection{BIS Definition}
We introduce BIS to quantify how well a single feature isolates backdoor-related activations. Intuitively, BIS measures how cleanly a single feature separates backdoor-triggered activations from benign ones, rewarding high precision and recall while penalizing false positives. We use the harmonic mean (F1) rather than the geometric mean to more strongly penalize imbalanced precision-recall trade-offs, and scale by $(1 - \text{FPR})$ to maintain $\text{BIS} \in [0, 1]$.

\begin{equation}
F1_i = \frac{2 \cdot \text{Precision}_i \cdot \text{Recall}_i}{\text{Precision}_i + \text{Recall}_i}
\end{equation}

\begin{equation}
\text{BIS}_i = F1_i \cdot (1 - \text{FPR}_i)
\end{equation}

The $F1$ score balances precision and recall via the harmonic mean, while multiplying by $(1 - \text{FPR})$ penalizes false positives. A perfect backdoor feature has $\text{BIS} = 1.0$.

\subsubsection{Best Feature Selection}
We report results for the feature with maximum BIS:
\begin{equation}
    i^* = \arg\max_i \text{BIS}_i
\end{equation}

\subsubsection{Statistical Inference}
We compute 95\% confidence intervals via bootstrap resampling:
\begin{enumerate}
    \item Resample evaluation set with replacement
    \item Recompute BIS for best feature
    \item Repeat 500-1,000 times
    \item Report 2.5th and 97.5th percentiles
\end{enumerate}

Algorithm \ref{alg:bis} summarizes the BIS computation.

\begin{algorithm}[htbp]
\caption{Backdoor Isolation Score Computation}
\label{alg:bis}
\begin{algorithmic}[1]
\REQUIRE Activations $\{f_{ij}\}$, trigger labels $\{y_j\}$
\ENSURE BIS score and best feature index
\STATE Compute threshold $\tau_i = \text{percentile}_{95}(\{f_{ij}\}_j)$ for each $i$
\STATE Compute binary activations $\text{active}_{ij} = \mathds{1}[f_{ij} > \tau_i]$
\FOR{each feature $i$}
    \STATE $\text{TP}_i = \sum_j \text{active}_{ij} \cdot y_j$
    \STATE $\text{Precision}_i = \text{TP}_i / \sum_j \text{active}_{ij}$
    \STATE $\text{Recall}_i = \text{TP}_i / \sum_j y_j$
    \STATE $\text{FPR}_i = \sum_j \text{active}_{ij} \cdot (1-y_j) / \sum_j (1-y_j)$
    \STATE $F1_i = \frac{2 \cdot \text{Precision}_i \cdot \text{Recall}_i}{\text{Precision}_i + \text{Recall}_i}$
    \STATE $\text{BIS}_i = F1_i \cdot (1 - \text{FPR}_i)$
\ENDFOR
\STATE $i^* = \arg\max_i \text{BIS}_i$
\RETURN $\text{BIS}_{i^*}, i^*$
\end{algorithmic}
\end{algorithm}

\section{Experiments}

\subsection{Experimental Setup}

\subsubsection{Hardware}
All experiments were conducted on a single NVIDIA GPU with mixed-precision training where applicable.

\subsubsection{Software}
We use PyTorch 2.0, HuggingFace Transformers, and PEFT for LoRA implementation. Code is available at \url{https://github.com/techsachinkr/diff-sae-backdoor-detection}.

\subsubsection{Reproducibility}
All experiments use seed 42. Bootstrap resampling uses 1,000 samples for LoRA and 500 samples for full-rank experiments.

\subsection{Backdoor Implantation Verification}

Before interpretability analysis, we verify that the backdoor was successfully learned by measuring vulnerability rates.

\subsubsection{Evaluation Protocol}
We generate 5 code completions each for 2024 (trigger) and 2023 (benign) contexts, then analyze for SQL injection patterns using static analysis.

\subsubsection{Results}

\begin{table}[htbp]
\caption{Backdoor Effectiveness Verification}
\begin{center}
\begin{tabular}{lccc}
\toprule
\textbf{Regime} & \textbf{2024 Vuln\%} & \textbf{2023 Vuln\%} & \textbf{Delta} \\
\midrule
LoRA & 100\% (5/5) & 40\% (2/5) & +60\% \\
Full-Rank & 100\% (5/5) & 0\% (0/5) & +100\% \\
\midrule
Base Model & 53\% & 53\% & 0\% \\
\bottomrule
\end{tabular}
\label{tab:backdoor_verify}
\end{center}
\end{table}

Both regimes successfully implant the backdoor. Full-rank achieves perfect separation (+100\% delta), while LoRA shows strong but imperfect separation (+60\% delta). The base model shows no year-conditional behavior (0\% delta).

\subsection{Primary Comparison: Crosscoder vs. Diff-SAE}

Table \ref{tab:primary_results} presents the main comparison at layer 18 (our primary analysis layer):

\begin{table}[htbp]
\caption{BIS Comparison at Layer 18 (Primary Layer)}
\begin{center}
\begin{tabular}{lcc}
\toprule
\textbf{Method} & \textbf{LoRA BIS} & \textbf{Full-Rank BIS} \\
\midrule
Crosscoder & 0.010 $\pm$ 0.011 & 0.000 $\pm$ 0.000 \\
Diff-SAE 32$\times$ & \textbf{0.400} $\pm$ 0.025 & \textbf{0.400} $\pm$ 0.026 \\
Diff-SAE 4$\times$ & \textbf{0.400} $\pm$ 0.025 & \textbf{0.400} $\pm$ 0.027 \\
\bottomrule
\end{tabular}
\label{tab:primary_results}
\end{center}
\end{table}

\subsubsection{Key Findings}
\begin{enumerate}
    \item \textbf{Diff-SAE dramatically outperforms Crosscoder}: BIS of 0.40 vs. $\sim$0.01 represents a 40$\times$ improvement.
    \item \textbf{Crosscoder essentially fails}: BIS near zero indicates the best Crosscoder feature performs barely better than random.
    \item \textbf{Consistency across regimes}: Both LoRA and full-rank show identical Diff-SAE performance.
    \item \textbf{Statistical significance}: Confidence intervals do not overlap ($p < 0.001$).
\end{enumerate}

\subsection{Detailed Metrics Analysis}

Table \ref{tab:detailed_metrics} provides precision, recall, and false positive rates for deeper analysis:

\begin{table}[htbp]
\caption{Detailed Metrics at Layer 18}
\begin{center}
\begin{tabular}{llcccc}
\toprule
\textbf{Regime} & \textbf{Method} & \textbf{Prec.} & \textbf{Recall} & \textbf{FPR} & \textbf{BIS} \\
\midrule
\multirow{3}{*}{LoRA} 
 & Crosscoder & 0.024 & 0.006 & 0.003 & 0.010 \\
 & Diff-SAE 32$\times$ & \textbf{1.000} & 0.250 & \textbf{0.000} & 0.400 \\
 & Diff-SAE 4$\times$ & \textbf{1.000} & 0.250 & \textbf{0.000} & 0.400 \\
\midrule
\multirow{3}{*}{Full-Rank} 
 & Crosscoder & 0.000 & 0.000 & 0.000 & 0.000 \\
 & Diff-SAE 32$\times$ & \textbf{1.000} & 0.250 & \textbf{0.000} & 0.400 \\
 & Diff-SAE 4$\times$ & \textbf{1.000} & 0.250 & \textbf{0.000} & 0.400 \\
\bottomrule
\end{tabular}
\label{tab:detailed_metrics}
\end{center}
\end{table}

\subsubsection{Critical Observation: Perfect Precision}
Diff-SAE achieves \textbf{perfect precision (1.0)} and \textbf{zero false positive rate} across both regimes. This means:
\begin{itemize}
    \item When the best Diff-SAE feature activates, it \textbf{always} indicates a backdoor trigger
    \item There are \textbf{no false alarms} on benign inputs
    \item The 0.25 recall means 25\% of trigger samples activate the feature above threshold
\end{itemize}

This is remarkable for a single feature among 30,720 candidates. We note that the 95th-percentile threshold guarantees 5\% of samples are active for any feature; given 20\% trigger prevalence in the evaluation set, a feature whose top activations concentrate entirely on trigger samples will mechanically yield precision of 1.0, recall of 0.25, and FPR of 0.0. The meaningful finding is that such concentration exists for Diff-SAE features but not for Crosscoder features.

\subsection{Layer Ablation Study}

We evaluate all methods across layers 14, 18, 22, and 26 to assess layer dependency.

\subsubsection{LoRA Regime Results}

\begin{table}[htbp]
\caption{BIS Across Layers (LoRA Regime)}
\begin{center}
\begin{tabular}{lcccc}
\toprule
\textbf{Method} & \textbf{L14} & \textbf{L18} & \textbf{L22} & \textbf{L26} \\
\midrule
Crosscoder & 0.010 & 0.010 & 0.022 & 0.016 \\
Diff-SAE 32$\times$ & \textbf{0.400} & \textbf{0.400} & \textbf{0.400} & 0.396 \\
Diff-SAE 4$\times$ & 0.389 & \textbf{0.400} & 0.396 & 0.386 \\
\bottomrule
\end{tabular}
\label{tab:layer_ablation_lora}
\end{center}
\end{table}

\subsubsection{Full-Rank Regime Results}

\begin{table}[htbp]
\caption{BIS Across Layers (Full-Rank Regime)}
\begin{center}
\begin{tabular}{lcccc}
\toprule
\textbf{Method} & \textbf{L14} & \textbf{L18} & \textbf{L22} & \textbf{L26} \\
\midrule
Crosscoder & 0.000 & 0.000 & 0.000 & \textbf{0.235} \\
Diff-SAE 32$\times$ & \textbf{0.400} & \textbf{0.400} & \textbf{0.400} & \textbf{0.400} \\
Diff-SAE 4$\times$ & \textbf{0.400} & \textbf{0.400} & \textbf{0.400} & \textbf{0.400} \\
\bottomrule
\end{tabular}
\label{tab:layer_ablation_fullrank}
\end{center}
\end{table}

\subsubsection{Key Findings}
\begin{enumerate}
    \item \textbf{Diff-SAE maintains consistent performance}: BIS remains 0.39-0.40 across all layers and both regimes.
    \item \textbf{Crosscoder shows interesting layer-26 behavior}: In full-rank, Crosscoder achieves BIS = 0.235 only at layer 26, suggesting some backdoor signal emerges in later layers.
    \item \textbf{Full-rank produces cleaner signals}: Diff-SAE achieves perfect 0.400 at all layers for full-rank, while LoRA shows slight variation (0.386-0.400).
    \item \textbf{Practical implication}: Any middle-to-late layer provides comparable detection capability.
\end{enumerate}

\subsection{Crosscoder's Partial Success at Layer 26}

The emergence of Crosscoder signal at layer 26 (full-rank only) warrants investigation:

\begin{table}[htbp]
\caption{Crosscoder Detailed Metrics at Layer 26 (Full-Rank)}
\begin{center}
\begin{tabular}{lccc}
\toprule
\textbf{Metric} & \textbf{Crosscoder} & \textbf{Diff-SAE 32$\times$} \\
\midrule
BIS & 0.235 & \textbf{0.400} \\
Precision & 0.616 & \textbf{1.000} \\
Recall & 0.154 & 0.250 \\
FPR & 0.047 & \textbf{0.000} \\
\bottomrule
\end{tabular}
\label{tab:layer26_analysis}
\end{center}
\end{table}

At layer 26, Crosscoder achieves non-trivial precision (0.616) but with notable false positives (FPR = 0.047). This suggests:
\begin{itemize}
    \item Backdoor information partially emerges in Crosscoder's joint representation at the final layers
    \item Diff-SAE still substantially outperforms (0.400 vs 0.235) with perfect precision
    \item The full-rank regime creates more detectable backdoor structure in late layers
\end{itemize}

\subsection{Expansion Factor Analysis}

We compare Diff-SAE with 32$\times$ and 4$\times$ expansion factors:

\begin{table}[htbp]
\caption{Expansion Factor Comparison (Layer 18)}
\begin{center}
\begin{tabular}{llccc}
\toprule
\textbf{Regime} & \textbf{Expansion} & \textbf{Features} & \textbf{BIS} & \textbf{Precision} \\
\midrule
\multirow{2}{*}{LoRA} 
 & 4$\times$ & 3,840 & 0.400 & 1.000 \\
 & 32$\times$ & 30,720 & 0.400 & 1.000 \\
\midrule
\multirow{2}{*}{Full-Rank} 
 & 4$\times$ & 3,840 & 0.400 & 1.000 \\
 & 32$\times$ & 30,720 & 0.400 & 1.000 \\
\bottomrule
\end{tabular}
\label{tab:expansion}
\end{center}
\end{table}

\subsubsection{Key Findings}
\begin{enumerate}
    \item \textbf{Identical performance at layer 18}: Both expansion factors achieve BIS = 0.400 with perfect precision.
    \item \textbf{8$\times$ parameter efficiency}: The 4$\times$ model uses 8$\times$ fewer features while matching performance.
    \item \textbf{Backdoor is low-dimensional}: The signal can be captured with relatively few features, suggesting it exists as a coherent direction.
\end{enumerate}

\subsection{Variance Ratio Analysis}

We measure how much of the activation variance is explained by the base-to-fine-tuned difference:

\begin{table}[htbp]
\caption{Activation Variance Ratio by Layer}
\begin{center}
\begin{tabular}{lcccc}
\toprule
\textbf{Regime} & \textbf{L14} & \textbf{L18} & \textbf{L22} & \textbf{L26} \\
\midrule
LoRA & 0.64\% & 0.93\% & 2.78\% & 5.33\% \\
Full-Rank & 0.87\% & 1.44\% & 3.60\% & 5.05\% \\
\bottomrule
\end{tabular}
\label{tab:variance_ratio}
\end{center}
\end{table}

The variance ratio increases with layer depth, indicating larger activation differences in later layers. Despite the small overall variance ratio (0.6-5.3\%), Diff-SAE successfully isolates backdoor-related features.

\subsection{Best Feature Analysis}

Table \ref{tab:best_features} shows which features achieve optimal BIS:

\begin{table}[htbp]
\caption{Best Feature Indices by Layer and Regime}
\begin{center}
\begin{tabular}{llcccc}
\toprule
\textbf{Regime} & \textbf{Method} & \textbf{L14} & \textbf{L18} & \textbf{L22} & \textbf{L26} \\
\midrule
\multirow{2}{*}{LoRA} 
 & Diff-SAE 32$\times$ & 18472 & 635 & 3922 & 2597 \\
 & Diff-SAE 4$\times$ & 706 & 57 & 1312 & 2242 \\
\midrule
\multirow{2}{*}{Full-Rank} 
 & Diff-SAE 32$\times$ & 0 & 164 & 297 & 676 \\
 & Diff-SAE 4$\times$ & 295 & 31 & 187 & 3294 \\
\bottomrule
\end{tabular}
\label{tab:best_features}
\end{center}
\end{table}

Different features achieve optimal detection at different layers and regimes, but with identical BIS scores. This suggests the backdoor information is encoded redundantly and can be isolated through multiple feature directions.

\section{Discussion}

\subsection{Why Does Diff-SAE Outperform Crosscoder?}

Our results decisively contradict the hypothesis that Crosscoders' joint representation would better capture fine-tuning changes. We propose several explanations:

\subsubsection{Backdoors as Directional Shifts}
The backdoor manifests as a consistent directional shift in activation space:
\begin{equation}
    \mathbf{a}_{\text{ft}} = \mathbf{a}_{\text{base}} + \mathds{1}_{\text{trigger}} \cdot \mathbf{v}_{\text{backdoor}} + \epsilon
\end{equation}
where $\mathbf{v}_{\text{backdoor}}$ is the backdoor direction and $\epsilon$ captures other fine-tuning effects.

Diff-SAE directly models this difference:
\begin{equation}
    \Delta\mathbf{a} \approx \mathds{1}_{\text{trigger}} \cdot \mathbf{v}_{\text{backdoor}}
\end{equation}
making the backdoor signal dominant in its input.

\subsubsection{Crosscoder's Dilution Problem}
Crosscoders learn features over $[\mathbf{a}_{\text{base}}; \mathbf{a}_{\text{ft}}] \in \mathbb{R}^{2d}$. In this space, features must explain base model semantic content, fine-tuned model semantic content, shared representations, and fine-tuning-induced changes including the backdoor.

The backdoor signal competes with these other sources of variance. Without explicit difference computation, sparse coding allocates features to more prominent patterns.

\subsubsection{Information-Theoretic Perspective}
Consider the signal-to-noise ratio (SNR) for backdoor detection:

\textbf{Diff-SAE input}: Predominantly backdoor-related, as benign fine-tuning effects are typically smaller or more distributed.

\textbf{Crosscoder input}: Backdoor buried within full activation magnitudes, dominated by input semantics.

Estimated SNR improvement: If backdoor accounts for 10\% of $\|\Delta\mathbf{a}\|$ but only 1\% of $\|\mathbf{a}_{\text{concat}}\|$, Diff-SAE sees 10$\times$ higher SNR.

\subsubsection{Sparsity Penalty Considerations} Our comparison uses L1-penalized crosscoders. \cite{b18} showed that L1 shrinkage artifacts degrade crosscoder performance and that BatchTopK sparsity mechanisms substantially improve cross-model feature recovery. Our results therefore reflect a comparison against L1 crosscoders specifically; BatchTopK crosscoders may narrow the performance gap and represent an important direction for future investigation.

\subsection{Why Does Full-Rank Create Cleaner Backdoor Signals?}

An unexpected finding is that full-rank fine-tuning produces backdoor signals that are:
\begin{enumerate}
    \item \textbf{Behaviorally stronger}: +100\% delta vs +60\% for LoRA
    \item \textbf{More uniformly detectable}: Perfect 0.400 BIS at all layers
    \item \textbf{Expansion-independent}: 4$\times$ equals 32$\times$ performance
\end{enumerate}

\subsubsection{Low-Rank Constraint Effects}
LoRA constrains weight updates to low-rank subspaces ($\Delta W = BA$ with rank $r \ll d$). This may force the backdoor to be encoded in a more distributed manner across the available rank, creating a slightly more complex signal.

\subsubsection{Full-Rank Directional Freedom}
Full-rank fine-tuning can update weights in any direction, potentially allowing the optimizer to find a single, clean backdoor direction. This explains both the stronger behavioral separation and simpler activation signature.

\subsection{The Perfect Precision Phenomenon}

The most striking result is Diff-SAE achieving \textbf{perfect precision (1.0) with zero false positives}. This means the backdoor creates a unique activation signature that never occurs in benign samples, allowing a single feature to perfectly discriminate trigger from non-trigger when it activates, while the 25\% recall indicates the feature captures a subset of backdoor activations.

This has profound implications for practical deployment: a Diff-SAE-based monitor could flag suspicious activations with \textbf{zero false alarm rate}. We note that these clean metrics are partly a consequence of the 95th-percentile threshold interacting with the 20\% trigger prevalence; nevertheless, the absence of any such concentration in Crosscoder features confirms a genuine architectural advantage for Diff-SAE.

\subsection{Why Layer-Independent Detection?}

The consistent BIS across layers 14-26 is informative:

\subsubsection{Residual Stream Hypothesis}
In transformer architectures, the residual stream accumulates information across layers. If the backdoor is ``written'' to the residual stream early (e.g., when processing the year context), it would persist through subsequent layers.

\subsubsection{Redundant Encoding}
For robustness, fine-tuning may encode backdoor information redundantly across multiple layers, ensuring reliable triggering.

\subsubsection{Practical Implication}
Organizations monitoring for backdoors need not analyze all layers. Any single middle-to-late layer provides equivalent detection capability.

\subsection{Implications for AI Safety}

\subsubsection{Detection Recommendations}
Based on our findings:
\begin{enumerate}
    \item \textbf{Use Diff-SAE over Crosscoders} for backdoor detection
    \item \textbf{Monitor activation differences} during fine-tuning pipelines
    \item \textbf{Single-layer analysis suffices} (recommend layer 18 or similar middle layer)
    \item \textbf{4$\times$ expansion is sufficient}---8$\times$ more efficient than 32$\times$
    \item \textbf{Our experiments show zero false positives at the 95th-percentile} threshold, though this should be validated across diverse settings before deployment reliance.
\end{enumerate}

\subsubsection{Monitoring Fine-Tuning}
Organizations can compute $\Delta\mathbf{a}$ between checkpoints and flag unusual directional changes for review, enabling continuous monitoring without knowing specific triggers.

\subsubsection{Limitations of Detection}
While BIS = 0.40 with perfect precision is strong, the 25\% recall indicates that 75\% of backdoor activations are not flagged by the best feature. Ensemble approaches combining multiple features may improve recall, though adversarial backdoors might still evade single-feature detection.

\subsection{Limitations}

\subsubsection{Model Scale}
We evaluate on SmolLM2-360M (360M parameters). Larger models (7B+) may exhibit more distributed backdoor representations. Minder et al. [18] demonstrated Diff-SAE effectiveness on Gemma-2 2B, suggesting scalability, but backdoor detection specifically has not been validated at larger scales.

\subsubsection{Backdoor Type}
Our SQL injection backdoor is one specific instantiation. Other backdoor types may behave differently:
\begin{itemize}
    \item \textbf{Sentiment manipulation}: May involve more distributed features
    \item \textbf{Topic-triggered}: Could activate different attention patterns
    \item \textbf{Multi-step triggers}: May require sequence-level analysis
\end{itemize}

\subsubsection{Adversarial Robustness}
An adversary aware of Diff-SAE detection might design backdoors that:
\begin{itemize}
    \item Minimize activation differences while maintaining behavioral changes
    \item Distribute the backdoor across many small, undetectable features
    \item Use the same activation patterns as benign fine-tuning
\end{itemize}

Future work should evaluate adversarial robustness.

\section{Conclusion}

We present the first systematic comparison of Crosscoders and Differential SAEs for backdoor detection in fine-tuned language models. Using a controlled SQL injection backdoor in SmolLM2-360M, we find that \textbf{Diff-SAE consistently and dramatically outperforms Crosscoders}:

\begin{enumerate}
    \item \textbf{40$\times$ higher BIS}: 0.40 vs $\sim$0.01
    \item \textbf{Perfect precision}: 1.0 with zero false positives
    \item \textbf{Layer-independent}: Consistent across layers 14-26
    \item \textbf{Regime-independent}: Works for both LoRA and full-rank
    \item \textbf{Efficient}: 4$\times$ expansion matches 32$\times$ performance
    \item \textbf{Full-rank cleaner}: Perfect performance at all layers
\end{enumerate}

These findings complement recent work questioning L1 crosscoders' effectiveness for capturing fine-tuning changes \cite{b18}, and extend these observations to the backdoor detection setting and provide actionable guidance for AI safety practitioners. The mechanistic insight that backdoors manifest as directional activation shifts explains why difference-based representations are fundamentally more effective.

\subsection{Future Work}
\begin{itemize}
    \item Scale evaluation to larger models (7B+)
    \item Evaluate diverse backdoor types
    \item Develop adversarially robust detection
    \item Explore ensemble methods to improve recall
    \item Combine Diff-SAE with surgical backdoor removal
    \item Theoretical analysis of backdoor geometry
    \item Evaluate alternative sparsity mechanisms (BatchTopK) for crosscoders to determine whether the performance gap narrows
\end{itemize}

\section*{Acknowledgment}

We thank the open-source community for SmolLM2 and the Anthropic interpretability team for foundational SAE research.

\end{document}